\documentclass[3p,times]{elsarticle}

\usepackage{ecrc}

\volume{00}

\firstpage{1}

\runauth{}

\jid{procs}

\usepackage{amssymb}

\usepackage[figuresright]{rotating}

\usepackage{times}
\usepackage{soul}
\usepackage{url}
\usepackage[hidelinks]{hyperref}
\usepackage[utf8]{inputenc}
\usepackage[small]{caption}
\usepackage{graphicx}
\usepackage{amsmath}
\usepackage{amsthm}
\usepackage{booktabs}
\usepackage{algorithm}
\usepackage{algorithmic}
\begin{document}

\begin{frontmatter}



\dochead{}

\title{Unpaired Referring Expression Grounding via Bidirectional Cross-Modal Matching}


\author{
	Hengcan Shi,
	Munawar Hayat,
	Jianfei Cai
}
\address{Monash University, Australia}

\begin{abstract}
Referring expression grounding is an important and challenging task in computer vision. To avoid the laborious annotation in conventional referring grounding, unpaired referring grounding is introduced, where the training data only contains a number of images and queries without correspondences. The few existing solutions to unpaired referring grounding are still preliminary, due to the challenges of learning vision-language correlation and lack of the top-down guidance with unpaired data. Existing works are only able to learn vision-language correlation by modality conversion, where critical information are lost. They also heavily rely on pre-extracted object proposals and thus cannot generate correct predictions with defective proposals.
In this paper, we propose a novel bidirectional cross-modal matching (BiCM) framework to address these challenges. Particularly, we design a query-aware attention map (QAM) module that introduces top-down perspective via generating query-specific visual attention maps to avoid the over-reliance on pre-extracted object proposals. A cross-modal object matching (COM) module is further introduced to predict the target objects from a bottom-up perspective. This module exploits the recently emerged image-text matching pretrained model, CLIP, to learn cross-modal correlation without modality conversion. The top-down and bottom-up predictions are then integrated via a similarity fusion (SF) module. We also propose a knowledge adaptation matching (KAM) module that leverages unpaired training data to adapt pretrained knowledge to the target dataset and task. Experiments show that our framework significantly outperforms previous works on three grounding datasets.
\end{abstract}


\end{frontmatter}


\section{Introduction}

Referring expression grounding, also called referring expression comprehension or natural language object localization, aims to localize objects from an image based on a language query. It serves as a fundamental step for many higher-level multi-modal tasks, such as image captioning \cite{duan2022position, tan2022acort, cao2020interactions}, cross-modal retrieval \cite{li2022image, liu2022featinter, dong2021multi} and cross-modal segmentation \cite{shi2018key, li2022cross, shi2020query}. 

\begin{figure*}
	\centerline{\includegraphics[scale=0.4]{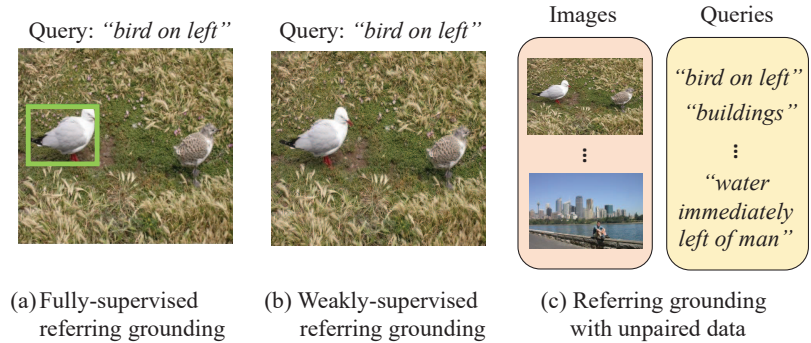}}
	\caption{Examples of training data in different referring grounding scenarios. (a) Fully-supervised grounding that provides images, corresponding queries and bounding box annotations for training. (b) Weakly-supervised grounding which contains training images and corresponding queries. (c) Unpaired referring grounding which only includes a set of images and a number of queries.}
	\label{introduction}
\end{figure*}

Fully-supervised referring grounding methods \cite{Hu2016Natural, mao2016generation, Zhang2017Discriminative, liu2017referring, qiu2020language, mu2021disentangled, huang2021look} have been well developed in recent years and achieve outstanding performance. They first use object detection networks to extract object proposals from the image, and then build scoring models to estimate the similarity between the language query and the extracted proposals. The desired object is the proposal with the highest similarity score. However, to train the scoring model, a large number of image-query pairs and corresponding bounding boxes need to be annotated, as shown in Fig. \ref{introduction} (a), which is a laborious and expensive process. To reduce annotation costs, weakly-supervised referring grounding methods \cite{xiao2017weakly, yeh2018unsupervised, chen2018knowledge, zhao2018weakly, liu2019adaptive, liu2019knowledge, gupta2020contrastive, zhang2020counterfactual} are proposed. As shown in Fig. \ref{introduction} (b), weakly-supervised grounding methods require images and corresponding queries for training. These methods use either cycle reconstruction or attention mechanism to avoid bounding box annotations. Nevertheless, labeling image-query pairs in the target dataset is still laborious and needs human experts.

To reduce the labor-intensive annotation, unpaired referring grounding has been attracting increasing research attention. The training set in this task only contains a number of images and some language queries, as depicted in Fig.~\ref{introduction}~(c). In real-world applications, images and queries can be separately collected from web (e.g., public images from Google and object descriptions from Wikipedia) without manual annotations, or randomly generated using some language templates or automatic language generation tools with little human efforts.
The key problem in this task is how to correlate vision and language without paired cross-modal training data.
Existing works solve this problem by extracting external knowledge from other tasks. Wang \emph{et al.} \cite{wang2019phrase} leverage knowledge from pretrained object detection models, which simultaneously generate object proposals and their class names. They calculate similarities between class names and the query in the textual space to select the target object. Parcalabescu \emph{et al.} \cite{parcalabescu2020exploring} further use pretrained scene graph models to convert some visual context into language words to better calculate similarities with the query. 

Despite significant progress made by these methods, a number of challenges remain unsolved. (1) There is a significant loss of information in the modality conversion in these methods. Either the class name or scene graph only contains limited features of objects, and critical clues such as color, position, posture, are not included in them. These features are very important for grounding, because queries often describe them. For example, in Fig. \ref{introduction}, it is hard to localize the ``bird on left'' by only using class names and scene graphs, since they do not contain the relative position information.
(2) Meanwhile, these methods heavily rely on pre-extracted object proposals. They will not be able to find the desired object if it is not in the pre-extracted proposals. Moreover, object proposals can only provide local visual information, where global visual information is ignored.
(3) These works only simply leverage knowledge from pretrained object detection models and scene graph generation models without any adaptation for the referring grounding task and the target data. Task and domain gaps between pre-learned knowledge and the target grounding dataset may decrease the accuracy.

On the other hand, a powerful image-text matching pretrained model, called CLIP (contrastive language-image pretraining) \cite{radford2021learning}, has been introduced recently, which consists of an image encoder and a text encoder. It was trained on millions of image-caption pairs by a contrastive loss. Due to its good generalization ability, CLIP has been used as external knowledge in several applications, such as image generation \cite{jalal2021fairness} and image classification \cite{cheng2021data}. This motivates us to consider: \textit{Can we exploit pretrained CLIP to address challenges in unpaired referring grounding?}

A straightforward way is to use CLIP image encoder to encode each proposal and CLIP text encoder to encode the query and then compare the encoded CLIP image and text directly. However, this only addresses the aforementioned first challenge since CLIP provides an aligned image-text embedding space. The other two challenges still remain. Therefore, in this paper, we propose a novel bidirectional cross-modal matching (BiCM) framework to solve all these challenges. Our BiCM framework contains four parts. In the first part, we introduce a query-aware attention map (QAM) module that generates predictions from a top-down perspective to reduce the over-reliance on object proposals. Specifically, it leverages CLIP to capture query-specific attention maps based on the global image and query, and generates predictions from the attention maps rather than object proposals. In the second part, we design a cross-modal object matching (COM) module to predict target objects from a bottom-up perspective. This module extracts object proposals and use the above straightforward way to directly compare visual object proposals with the textual query to avoid information loss in modality conversion. Thirdly, a similarity fusion (SF) module is further designed to fuse the bottom-up and top-down matching results. In the fourth part, we propose a learnable knowledge adaptation matching (KAM) module, which adapts pretrained CLIP knowledge to the target grounding data to solve domain and task gaps. Extensive experiments on Flickr30K Entities \cite{plummer2015flickr30k}, ReferItGame \cite{Kazemzadeh2014ReferItGame} and Google-Ref dataset \cite{mao2016generation} datasets demonstrate the effectiveness of our framework.

Our major contributions can be summarized as follows. 
\begin{enumerate}
	\item We propose a novel bidirectional cross-modal matching (BiCM) framework for unpaired referring grounding, where we design four components (QAM, COM, SF and KAM) to predict grounding results from both top-down and bottom-up perspectives and allow target-specific knowledge adaptation. 
	\item To the best of our knowledge, this is the first study to exploit CLIP knowledge for unpaired referring grounding. We explore CLIP feature space for cross-modal matching and propose a QAM module to extract query-aware visual attention maps from CLIP.
	\item Extensive experimental results show that our proposed framework obtains significant improvements on three popular referring grounding datasets.
\end{enumerate}

\section{Related Work}
\textbf{Fully-supervised referring grounding.} Early referring expression grounding works \cite{Hu2016Natural, mao2016generation, Zhang2017Discriminative} first used object detection models to extract a number of candidate bounding boxes from the image. Then, cross-modal classifiers were trained to score each candidate bounding box based on its features and the language query features. Finally, The bounding box with the highest score was chosen as the result. These works can only select bounding boxes from pre-extracted candidates and cannot adjust bounding boxes. To adjust them, Yeh \emph{et al.} \cite{yeh2018interpretable} leveraged bounding boxes, segmentation masks and the language query to generate score maps and then employed efficient subwindow search \cite{lampert2009efficient} to predict the desired box. Nagaraja \emph{et al.} \cite{Nagaraja2016Modeling}, Yu \emph{et al.} \cite{Yu2016Modeling} and Zhang \emph{et al.} \cite{zhang2018grounding} incorporated context models to learn global object relationships to improve the grounding accuracy. Liu \emph{et al.} \cite{liu2017referring} modeled attributes of visual objects and their paired descriptions to enhance the vision-language matching. Qiu \emph{et al.} \cite{qiu2020language} designed language-aware deformable convolutions to allow fine-grained cross-model matching. Attention models \cite{yu2018mattnet, anayurt2019searching, yang2019dynamic} are also exploited to suppress noises in the image and language query, and thus can extract more discriminative visual and textual features. To further boost the grounding performance, some methods incorporated extra supervisions from other tasks such as image caption \cite{yu2017joint, rohrbach2016grounding} and referring segmentation \cite{luo2020multi}. Moreover, multi-head-self-attention-based Transformers \cite{kamath2021mdetr, deng2021transvg} are built to end-to-end grounding. A major limitation of these methods is the required human annotation effort, which is very expensive (specially for large-scale datasets with massive language queries and bounding boxes).

\textbf{Weakly-supervised referring grounding.} Weakly supervised methods have recently attracted increasing research interests \cite{xiao2017weakly, yeh2018unsupervised, chen2018knowledge, zhao2018weakly, liu2019adaptive, liu2019knowledge, gupta2020contrastive, zhang2020counterfactual, liu2021relation}. They do not need to annotate bounding boxes and only require image-query pairs. With image-query training pairs, Xiao \emph{et al.} \cite{xiao2017weakly} trained a deep neutral network to match the entire image with the language query, and generated the target bounding box from highlighted parts in feature maps in this network. Yeh \emph{et al.} \cite{yeh2018unsupervised} leveraged the image-query-pair training data to make co-occurrence statistics between bounding box label names and words in queries. Co-occurrence statistics results were used to predict bounding boxes in the testing set. Several works \cite{chen2018knowledge, zhao2018weakly, liu2019adaptive, liu2019knowledge} used cycle reconstruction mechanism to allow weakly-supervised training. Their network architectures are similar to fully-supervised referring grounding networks. Nevertheless, due to no bounding box ground-truth during training, these methods reconstructed the language query and the image from the predicted bounding box, and calculated similarities between reconstructed ones and input ones as loss functions. Besides image-query matching, co-occurrence statistics and cycle reconstruction, contrastive learning \cite{gupta2020contrastive, zhang2020counterfactual} was also proposed for weakly-supervised training. In contrastive learning, networks are trained by comparing positive and negative samples. Gupta \emph{et al.} \cite{gupta2020contrastive} regarded the original image-query pair as the positive sample and generated two negative samples. The first is the language query with another image, while the second is the original image with a changed query. Zhang \emph{et al.} \cite{zhang2020counterfactual} proposed to generate positive and negative training data though three counterfactual transformations, containing feature-level, interaction-level and relation-level transformations.  Visual attributes \cite{liu2019adaptive}, attention models \cite{liu2019adaptive, liu2019knowledge, gupta2020contrastive} and language parsing \cite{xiao2017weakly, liu2019knowledge} were also employed to improve the grounding accuracy. Although these works avoid annotation costs of bounding boxes, they still need to label image-query pairs. For large-scale datasets, annotating image-query pairs is still costly and needs human experts.

\textbf{Unpaired referring grounding.} To further reduce annotation costs and avoid the requirement of human annotators, referring grounding with unpaired data \cite{wang2019phrase, parcalabescu2020exploring} is a promising learning paradigm. In this task, there is neither phrase-image pairs nor bounding boxes in the training set. Wang \emph{et al.} \cite{wang2019phrase} used object detection models trained on visual detection datasets to predict candidate bounding boxes and their class names. Next, they leveraged textual datasets to train a text encoder and extracted features of the language query and box class names. Finally, they select the desired bounding box by comparing each box class name features with the query features. Parcalabescu \emph{et al.} \cite{parcalabescu2020exploring} incorporated visual and textual context to Wang \emph{et al.}'s method. For vision, they used pre-trained scene graph generation methods to extract scene graph of the input image as visual context. To model textual context, their text encoder was trained on hierarchy datasets which label the hypernymy, hyponymy and synonym for each word. These captured visual and textual context are used to enrich the features of each box class name. However, these methods only simply employ pre-learned knowledge without any adaptation, and there are many information losses in modal conversions, which degrades their performance. In addition, they highly rely on pre-extracted proposals and class names, and ignore task and domain gaps between pretrained models and target datasets.

\section{Proposed Method}
\begin{figure*}[h]
	\centerline{\includegraphics[scale=0.4]{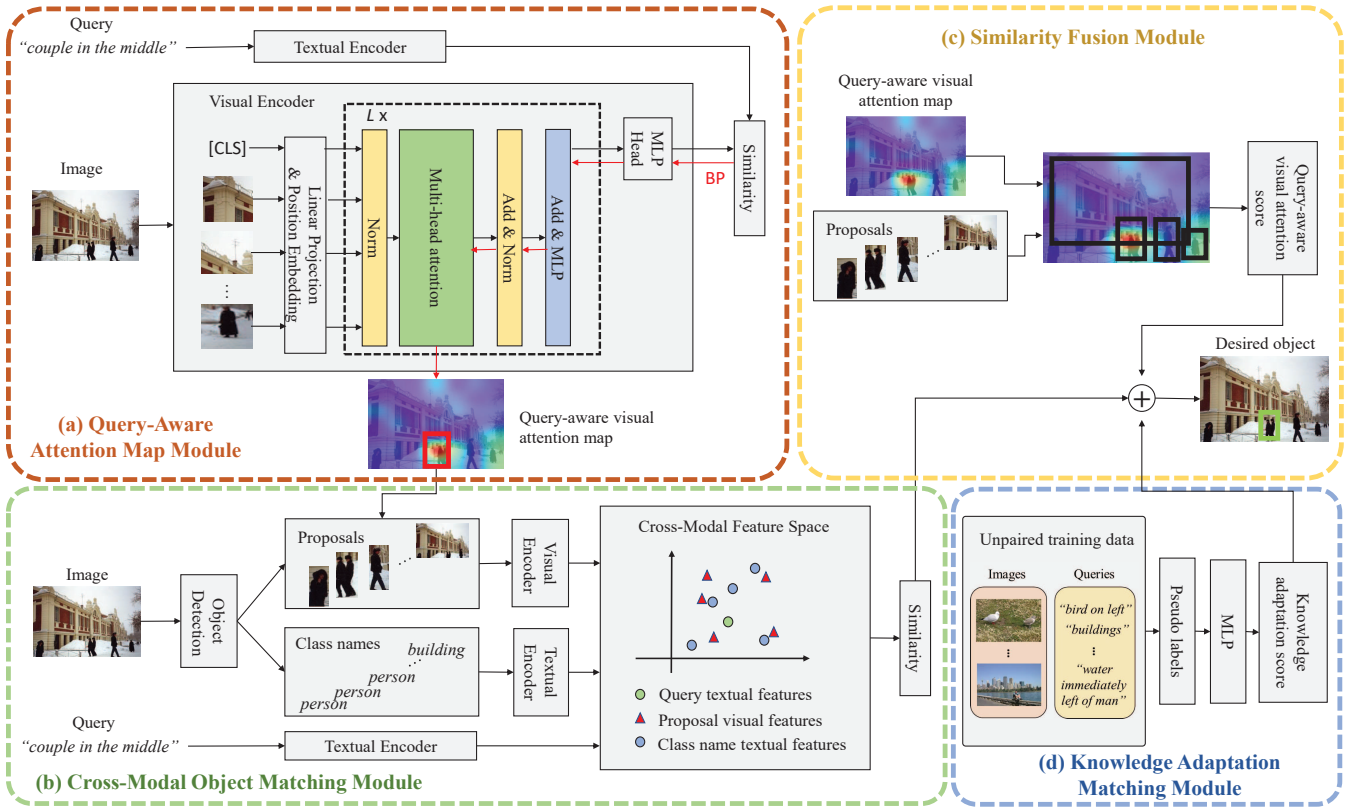}}
	\caption{Illustration of BiCM. (a) A top-down QAM module extracts a query-aware visual attention map by back propagation (BP) and predicts a grounding result (the red box). (b) A bottom-up COM module selects the desired object from object proposals. The result from QAM is added, to avoid the over-reliance on pre-extracted proposals. (c) An SF module fuses the predictions from QAM and COM. (d) A KAM module leverages unpaired training data to adapt pretrained knowledge to the target grounding dataset to further improve the accuracy.}
	\label{fig_method}
\end{figure*}

\subsection{Problem Definition and Method Overview}
The inputs of referring grounding are an image $I $ and a language query $Q$. Referring grounding expects to output a bounding box $P$ of the object described by the query. In this paper, we propose a bidirectional cross-modal matching (BiCM) framework to predict the bounding box.

Our BiCM framework contains four components as shown in Fig.~\ref{fig_method}: (a) a query-aware attention map (QAM) module that generates attention maps and predicts a candidate bounding box in a top-down manner; (b) a cross-modal object matching (COM) module that leverages CLIP feature space to select bounding boxes from object proposals in a bottom-up manner; (c) a similarity fusion (SF) module that integrates the top-down and bottom-up results; (d) and a knowledge adaptation matching (KAM) module that adapts CLIP knowledge to our target grounding data. We introduce the details of each component below.

\subsection{Query-Aware Attention Map Module}\label{GridCLIP}
The top-down QAM module aims to generate the grounding result from the global image and query. Inspired by Grad-CAM \cite{selvaraju2017grad} in CNN, we design a back-propagation-based mechanism to extract query-aware visual attention maps in Transformers to achieve this goal. 

As shown in Fig. \ref{fig_method} (a), we first input the image $I$ into the pre-trained CLIP visual encoder (ViT \cite{dosovitskiy2020image}) and output the feature vector $\mathbf{fi}$.
In ViT, the input image is divided into $U$ patches (tokens) and a special [CLS] token is added to generate $\mathbf{fi}$. Visual attention is extracted from the last block in ViT as follows:
\begin{equation}
	att_{h, u} = Softmax(\frac{\mathbf{q}_{h} \mathbf{k}_{h, u}^{T}}{\sqrt{d}})
\end{equation}
where $h = 1, ..., H$ and $H$ is the number of heads in the last block. $\mathbf{q}_{h}$ is the attention ``query'' of the [CLS] token in the $h$-th head, and $\mathbf{k}_{h, u}$ represents the attention ``key'' of the $u$-th image patch. $d$ is a scaling factor. $att_{h, u}$ is the attention score of the $u$-th image patch in the $h$-th head, which means the importance of this image patch for the final output. It can be used to highlight visually salient regions in the image. However, this attention score cannot find out the regions corresponding to the language query $Q$.

To generate query-aware attention scores, we leverage the pre-trained CLIP textual encoder (GPT-2 \cite{radford2019language}) to encode the feature vector $\mathbf{fq}$ of the input query $Q$. Since the image feature vector $\mathbf{fi}$ and the query feature vector $\mathbf{fq}$ are embedded into the same feature space, we calculate their cosine similarity $S^{IQ}$ as
\begin{equation}
	S^{IQ} = \frac{<\mathbf{fi}, \mathbf{fq}>}{\Vert \mathbf{fi} \Vert \Vert \mathbf{fq} \Vert}
\end{equation}
where $<,>$ represents inner product.

We then take the image-query similarity $S^{IQ}$ as a loss and propagate it back into the visual encoder:
\begin{equation}
	\alpha_{h, u} = \frac{\partial S^{IQ}}{\partial att_{h, u}}
\end{equation}
where $\alpha_{h, u}$ is the gradient on $att_{h, u}$, which can be regarded as the importance of this attention score for the image-query similarity $S^{IQ}$. Thus, we use it to weight the attention score to find out the most important image patches for the language query as
\begin{equation}
	\widetilde{att}_{h, u} = att_{h, u} \times ReLU(\alpha_{h, u})
\end{equation}
where $\widetilde{att}_{h, u}$ is the weighted attention score, and ReLU function is used to filter out the negative importance scores. We average scores in all heads as our final query-aware visual attention score $qa_{u}$:
\begin{equation}
	a_{u} = \frac{1}{H}\sum_{h=1}^{H}\widetilde{att}_{h, u}.
\end{equation}
Scores $a_{u}$ for all image patches construct a query-aware visual attention map $A$. We finally upsample $A$ into the original image size by bilinear interpolation and use min-max normalization to normalize it. By empirically setting a threshold $Thr_{a}$, we can select out the desired object and generate the bounding box $P_{t}$.

\subsection{Cross-Modal Object Matching Module}\label{first_matching}
Our QAM module is able to localize objects from the top-down perspective. Nevertheless, it cannot always capture compact object boundaries. Thus, we further introduce a bottom-up method to identify objects from pre-extracted object proposals.

In particular, an object detection model such as Faster RCNN \cite{ren2015faster} is first used to generate object proposals $\{P_{r} \}_{r=1}^{N_{r}}$ from the input image $I$, where $N_{r}$ denotes the number of proposals in the image. Then, we leverage the CLIP visual encoder to extract visual feature vector $\mathbf{fp}_{r}$ of each proposal. After that, we compute the similarity $S^{PQ}_{r}$ between the query and each proposal as
\begin{equation}
	S^{PQ}_{r} = \frac{<\mathbf{fp}_{r}, \mathbf{fq}>}{\Vert \mathbf{fp}_{r} \Vert \Vert \mathbf{fq} \Vert}.
\end{equation}
This similarity directly compares the visual features of objects with the language query, and thus avoids information loss caused by modality conversion.

In addition, previous works \cite{wang2019phrase,parcalabescu2020exploring} indicate that class names of object proposals are also important information. Therefore, we further calculate the similarity between the query $Q$ and each proposal class name $C_{r}$. We use the CLIP textual encoder to encode $C_{r}$ into the feature vector $\mathbf{fc}_{r}$ and then compute the cosine similarity:
\begin{equation}
	S^{CQ}_{r} = \frac{<\mathbf{fc}_{r}, \mathbf{fq}>}{\Vert \mathbf{fc}_{r} \Vert \Vert \mathbf{fq} \Vert}
\end{equation} 

The final bottom-up similarity $S^{BU}_{r}$ between each proposal and the query is defined as the sum of the two similarities:
\begin{equation}
	S^{BU}_{r} = S^{PQ}_{r} + S^{CQ}_{r}.
\end{equation} 
The proposal $P_{r}$ with the highest similarity could be selected as the desired object. 

Benefiting from the pretrained object detection model, our bottom-up COM module can extract compact object boundaries. However, it cannot predict the desired object if the pretrained detection model misses the object. To avoid this, we add our top-down prediction $P_{t}$ to the set of object proposals $\{P_{r} \}_{r=1}^{N_{r}}$ and select the desired object from the combined set $\{P_{1}, ..., P_{N_{r}}, P_{t} \}$. We take the proposal class name $C_{r}$ which has the highest similarity $S^{CQ}_{r}$ with the query as the class name of $P_{t}$.

\subsection{Similarity Fusion Module}
The SF module is to integrate the similarity scores from the top-down QAM module and the bottom-up COM module, as shown in Fig. \ref{fig_method} (c). For each object proposal in $\{P_{1}, ..., P_{R}, P_{t}\}$, we generate its top-down similarity based on the query-aware visual attention map $A$ as
\begin{equation}
	S_{r}^{TD} = \frac{1}{N_v}\sum_{v=1}^{N_{v}} a_{v}
\end{equation}
where $N_{v}$ is the number of pixels in this proposal and $a_{v}$ is the attention score of each pixel $v$ in $A$.

We then fuse the top-down and bottom-up scores as follows:
\begin{equation}
	S_{r} = S_{r}^{BU} + \lambda_{t} S_{r}^{TD}
\end{equation}
where $\lambda_{t}$ is the weight to trade off the two scores. The final result is the proposal whose fused score is the highest.

\subsection{Knowledge Adaptation Matching Module} \label{sec_pseudo_label}
The purpose of this KAM module is to generate pseudo labels from unpaired training data and train a lightweight network to adapt CLIP knowledge to the target dataset and the grounding task. Specifically, given the unpaired image and query sets, we leverage CLIP visual and textual encoders to obtain the image and query features, respectively. For each image, we find out the query with the highest similarity, and treat the image and the query as a pseudo image-query pair. Then, for each image-query pair, we use the above three components to predict a bounding box. If its fused similarity score $S_r$ is higher than a threshold $Thr_{k}$, we choose the bounding box as a pseudo label or pseudo ground-truth. 

\begin{figure}[h]
	\centerline{\includegraphics[scale=0.42]{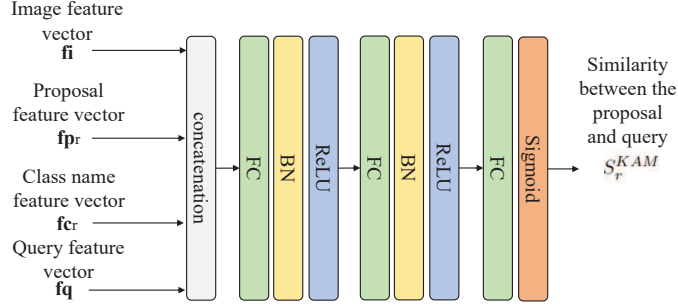}}
	\caption{Illustration of the MLP in KAM. ``'FC' means the fully connected layer. ``BN'' represents the batch normalization layer. ``ReLU'' and ``Sigmoid'' are ReLU activation and Sigmoid normalization, respectively.}
	\label{fig_KAM}
\end{figure}

After constructing pseudo labels, we then train a simple MLP (Multi-layer Perception) network, which takes the image features $\mathbf{fi}$, visual object proposal features $\mathbf{fp}_{r}$, object class name features $\mathbf{fc}_{r}$ and query features ${\mathbf{fq}}$ as inputs, and outputs another similarity score $S_{r}^{KAM}$. As shown in Fig. \ref{fig_KAM}, our MLP in KAM consists of three fully-connected layers with batch normalization and ReLU activation. The first layer is used to fuse input features, the second layer is to transform the fused features and the final layer outputs the score $S_{r}^{KAM}$ which is normalized by a Sigmoid function. We adopt the loss in fully-supervised grounding \cite{Hu2016Natural} to train the MLP.

During inference, the score $S_{r}^{KAM}$ can be added to $S_{r}$ to select the target object:
\begin{equation}
	Sim_{r} = S_{r}^{BU} + \lambda_{t} S_{r}^{TD} + \lambda_{k} S_{r}^{KAM}
\end{equation}
where $\lambda_{k}$ is the weight of $S_{r}^{KAM}$.

\section{Experiments}
\subsection{Experimental Setting}
We evaluate our method on three referring grounding datasets, including the Flickr30K Entities \cite{plummer2015flickr30k}, ReferItGame \cite{Kazemzadeh2014ReferItGame}, Google-Ref \cite{mao2016generation} datasets.

The Flickr30K Entities dataset \cite{plummer2015flickr30k} is a phrase grounding dataset. There are 31,783 images and 158,915 descriptions (five descriptions per image) in this dataset. 513,644 phrases in these descriptions describe 275,775 bounding boxes in images. Many phrases (e.g., \emph{several people}) describe multiple bounding boxes in an image. In this case, following previous methods \cite{wang2019phrase, parcalabescu2020exploring}, we merge these bounding boxes and use the union region as the ground-truth. The entire dataset is split into training, validation and testing sets, containing 29783, 1000 and 1000 images, respectively. We use unpaired data in training and validation sets to train our model, and evaluate it on the testing set. The testing set comprises 14,481 phrases for 1,000 images.

The ReferItGame dataset \cite{Kazemzadeh2014ReferItGame}, also known as RefCLEF, contains 20,000 images, 9,000 for training, 1,000 for validation and 10,000 for testing. 130,525 phrase expressions describe 96,654 objects in these images. It is a more challenging dataset, because phrase lengths in this dataset are often longer than that in the Flickr30K Entities dataset. Different from the Flickr30K Entities dataset, every phrase in the ReferItGame dataset only describe one object bounding box. We also leverage unpaired data in training and validation sets to train, and use the testing set to estimate the accuracy of our model.

The Google-Ref dataset \cite{mao2016generation} collects 26,711 images from the MS COCO dataset \cite{lin2014microsoft}. There are 54,822 objects and 104,560 referring expressions, which are divided into training and validation, including 44,822 and 5,000 objects, respectively. We use the training set for training, and verify our model on the validation set.

\textbf{Metrics.} We adapt the grounding accuracy to estimate our grounding framework, which is percentage of predictions whose IoU with ground truth is higher than 0.5.

\textbf{Implementation Details.} We use Faster RCNN \cite{ren2015faster} pretrained on Visual Genome \cite{krishna2017visual} to extract 100 object proposals for each image, and encode 512-dimension visual and textual features. Thresholds $Thr_{a}$ and $Thr_{k}$ are set to 0.7 and 0.9, respectively. We set $\lambda_{t}$ and $\lambda_{k}$ to 1000 and 1, to make the three scores have similar orders of magnitude. Our MLP is trained on one Nvidia RTX 3090 GPU for 50 epochs. Adam optimizer is used for training and the base learning rate is set to 0.0001.  On the Flickr30K Entities dataset, because some queries refer to multiple bounding boxes in an image, all methods including ours merge multiple high-score bounding boxes as final results. We merge bounding boxes whose similarities are higher than the average similarity. On other datasets, a query only describes one object in an image. To fairly compared with previous work \cite{wang2019phrase}, we select the largest bounding box from above-average-similarity bounding boxes as our prediction. 

\subsection{Results and Comparisons}
Table \ref{tab_Grounding accuracy on Flickr30K Entities} shows results of our and other state-of-the-art methods on the Flickr30K Entities dataset. For a fair comparison, all unpaired-data methods use Faster RCNN pretrained on Visual Genome \cite{krishna2017visual} to extract proposals. Moreover, in \cite{parcalabescu2020exploring}, results of \cite{wang2019phrase} and \cite{parcalabescu2020exploring} on a non-standard testing set with 16,576 phrases are reported. We reproduce these methods on the standard testing set with 14,481 queries. Compared with Wang \emph{et al.}'s method \cite{wang2019phrase} which only uses class name information, our method achieves improvements by 8.21\%. Parcalabescu \emph{et al.} \cite{parcalabescu2020exploring} employ scene graphs to improve referring grounding in their method. Our method outperforms their method by 6.55\%.

Results on the ReferItGame dataset are shown in Table \ref{tab_Grounding accuracy on ReferItGame}. It can be observed that our method outperforms \cite{wang2019phrase} and \cite{parcalabescu2020exploring} by 15.40\% and 12.88\%, respectively. Results on the Google-Ref dataset are shown in Table \ref{tab_Google_ref}, where our method also significantly outperforms previous methods on all sets. These results demonstrate the effectiveness of our BiCM framework.

In Table \ref{tab_Grounding accuracy on Flickr30K Entities}, Table \ref{tab_Grounding accuracy on ReferItGame} and Table \ref{tab_Google_ref}, we report results from some fully- and weakly-supervised methods as references. Our method and previous unpaired methods \cite{wang2019phrase,parcalabescu2020exploring} outperform many weakly-supervised methods. A reason is that we introduce external knowledge. Our method also shows competitive accuracy against some fully-supervised methods, such as \cite{dogan2019neural} and \cite{plummer2017phrase}. 

\begin{figure*}
	\centerline{\includegraphics[scale=0.42]{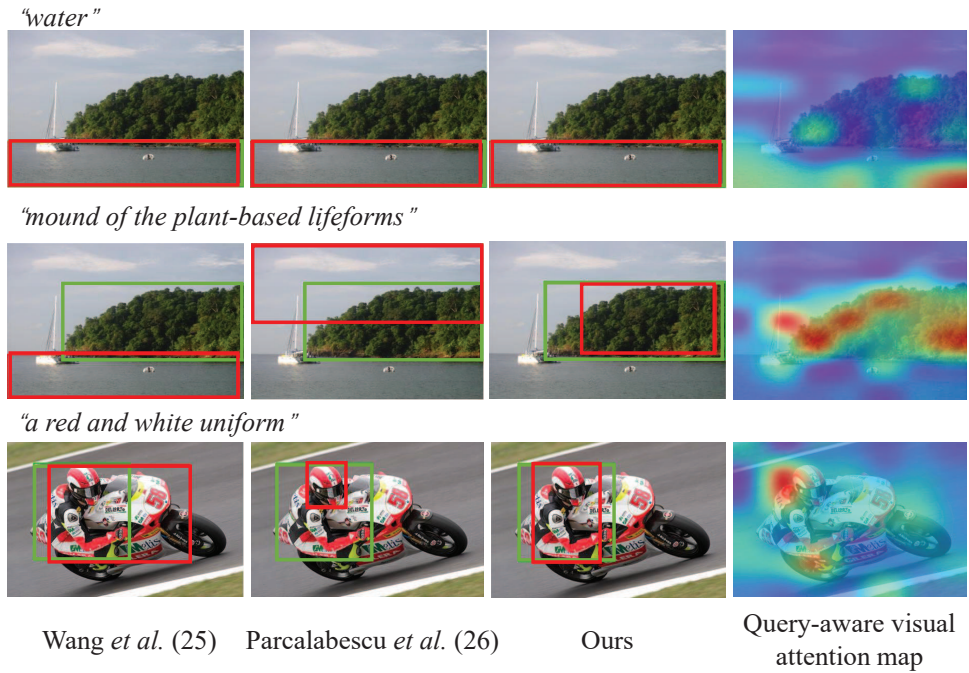}}
	\caption{Grounding results of different methods, and our query-aware visual attention maps on ReferItGame test set (the first and second rows) and the Flickr30K Entities test set (the third row). Red boxes denote predictions and green boxes are ground truths. }
	\label{fig_results_referit}
\end{figure*}

\begin{table}
	\centering
	\caption{Grounding accuracy on the Flickr30K Entities test. ``*'' means results estimated on a non-standard testing set with 16,576 phrases.}
	\scalebox{0.8}{
		\begin{tabular}{lc}
			\toprule
			Method                                                  & Accuracy (\%)  \\
			\midrule
			\emph{Fully-supervised} \\
			Rohrbach \emph{et al.} \cite{rohrbach2016grounding}                            & 47.81    \\
			Plummer \emph{et al.} \cite{plummer2017phrase}                                & 55.85    \\
			Dogan \emph{et al.} \cite{dogan2019neural}                                & 61.60    \\
			Yang \emph{et al.} \cite{yang2020propagating}                                & 69.53    \\   
			Liu \emph{et al.} \cite{liu2020learning}                                   & 76.74      \\
			Deng \emph{et al.} \cite{deng2021transvg}                                & 78.47    \\
			Mu \emph{et al.} \cite{mu2021disentangled}                                & \textbf{78.73}    \\
			\hline
			\emph{Weakly-supervised} \\
			Rohrbach \emph{et al.} \cite{rohrbach2016grounding}                            & 28.94    \\
			Zhao \emph{et al.} \cite{zhao2018weakly}                                   & 33.10    \\
			Yeh \emph{et al.} \cite{yeh2018unsupervised}                              & 36.93    \\
			Chen \emph{et al.} \cite{chen2018knowledge}                                & 38.71    \\
			Gupta \emph{et al.} \cite{gupta2020contrastive}                             & 51.67    \\
			Wang \emph{et al.} \cite{wang2021improving}                             & 53.10    \\
			Liu \emph{et al.} \cite{liu2021relation}                                  & \textbf{59.27}    \\
			\hline
			\emph{Unpaired Data} \\
			Wang \emph{et al.} \cite{wang2019phrase}                                   & 53.25      \\
			Wang \emph{et al.} \cite{wang2019phrase}*                                  & 56.30      \\
			Parcalabescu \emph{et al.} \cite{parcalabescu2020exploring}                        & 54.91   \\
			Parcalabescu \emph{et al.} \cite{parcalabescu2020exploring}*                       & 57.08   \\
			Ours                                                    &\textbf{61.46} \\
			\bottomrule
	\end{tabular}}
	\label{tab_Grounding accuracy on Flickr30K Entities}
\end{table}

\begin{table}
	\centering
	\caption{Grounding accuracy on the ReferItGame test.}
	\scalebox{0.8}{
		\begin{tabular}{lc}
			\toprule
			Method                                                  & Accuracy (\%)  \\
			\midrule
			\emph{Fully-supervised} \\
			Hu \emph{et al.} \cite{Hu2016Natural}                                    & 17.93    \\			
			Rohrbach \emph{et al.} \cite{rohrbach2016grounding}                            & 26.93    \\
			Zhang \emph{et al.} \cite{zhang2018grounding}                                & 31.13    \\ 
			Plummer \emph{et al.} \cite{plummer2018conditional}                            & 34.15    \\ 
			Bajaj \emph{et al.} \cite{bajaj2019g3raphground}                            & 44.91    \\
			Mu \emph{et al.} \cite{mu2021disentangled}                                & 65.15    \\
			Huang \emph{et al.} \cite{huang2021look}                                     & 67.47 \\
			Deng \emph{et al.} \cite{deng2021transvg}                                & \textbf{69.76}    \\
			\hline
			\emph{Weakly-supervised} \\
			Rohrbach \emph{et al.} \cite{rohrbach2016grounding}                            & 10.70    \\
			Zhao \emph{et al.} \cite{zhao2018weakly}                                   & 13.61    \\
			Chen \emph{et al.} \cite{chen2018knowledge}                                & 15.83    \\
			Yeh \emph{et al.} \cite{yeh2018unsupervised}                              & 20.91    \\
			Liu \emph{et al.} \cite{liu2019adaptive}                             & 26.19    \\
			Liu \emph{et al.} \cite{liu2021relation}                                  & 37.68    \\
			Wang \emph{et al.} \cite{wang2021improving}                             & \textbf{38.39}    \\
			\hline
			\emph{Unpaired Data} \\
			Wang \emph{et al.} \cite{wang2019phrase}                                   & 27.56   \\
			Parcalabescu \emph{et al.} \cite{parcalabescu2020exploring}                        & 30.08   \\
			Ours                                                    &\textbf{42.96} \\
			\bottomrule
	\end{tabular}}
	\label{tab_Grounding accuracy on ReferItGame}
\end{table}

\begin{table}
	\centering
	\caption{Grounding accuracy on the Google validation.}
	\scalebox{0.8}{
		\begin{tabular}{lc}
			\toprule
			Method                              &  Accuracy (\%) \\
			\midrule
			\emph{Fully-supervised} \\
			Mao \emph{et al.} \cite{mao2016generation}  & 44.50 \\
			Yu \emph{et al.} \cite{yu2018mattnet}       &66.58 \\
			Huang \emph{et al.} \cite{huang2021look}    &62.70 \\
			Deng \emph{et al.} \cite{deng2021transvg}   &\textbf{67.02}\\
			\emph{Fully-supervised} \\
			\hline
			\emph{Weakly-supervised} \\
			Liu \emph{et al.} \cite{liu2019knowledge}  & 38.37 \\
			Liu \emph{et al.} \cite{liu2019adaptive}   &\textbf{39.62} \\ 
			\hline
			\emph{Unpaired Data} \\
			Wang \emph{et al.} \cite{wang2019phrase}                     & 37.76  \\
			Parcalabescu \emph{et al.} \cite{parcalabescu2020exploring}  & 43.93 \\
			Ours                                & \textbf{52.85}\\
			\bottomrule
	\end{tabular}}
	\label{tab_Google_ref}
\end{table}

\begin{figure*}
	\centerline{\includegraphics[scale=0.4]{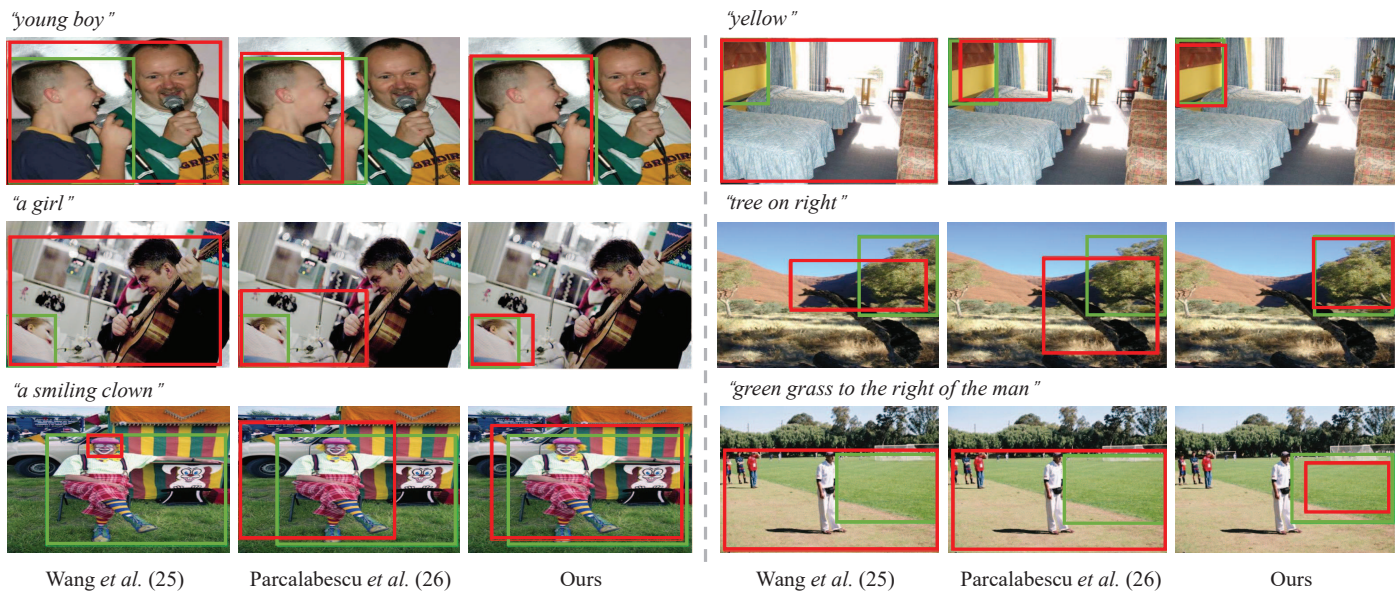}}
	\caption{Grounding results on Flickr30K Entities test (left three columns) and ReferItGame test (right three columns). Red boxes denote predictions while green boxes are ground truths. }
	\label{fig_results}
\end{figure*}

We visualize some grounding results in Fig.~\ref{fig_results_referit}. It is seen that while previous methods do well at finding objects when the query is a noun (the first row in Fig.~\ref{fig_results_referit}), they mis-localize many objects for relatively complex queries (e.g., the second and third rows in Fig.~\ref{fig_results_referit}. The reason is that previous technologies only use class names and scene graphs to compare with queries, while some key information described in these queries are not contained in object class names, such as color and position. Different from previous methods, our method leverages bidirectional matching and knowledge adaptation, and thus avoids these grounding errors. Fig. \ref{fig_results} shows more qualitative grounding results on the Flickr30K Entities and ReferItGame datasets. Compared with previous methods \cite{wang2019phrase,parcalabescu2020exploring}, our method finds out the desired objects more accurately.

\subsection{Discussions}
\begin{table}
	\centering
	\caption{The effects of the main components on Flickr30K Entities test.}
	\scalebox{1.0}{
		\begin{tabular}{lc}
			\toprule
			Method                           & Accuracy (\%)  \\
			\midrule
			Baseline \cite{wang2019phrase}   & 53.25 \\
			QAM                              & 30.86\\
			COM                              & 57.60 \\
			QAM+COM                          & 58.43\\
			QAM+COM+SF      & 60.15\\
			QAM+COM+SF+KAM  & 61.46\\
			\bottomrule
	\end{tabular}}
	\label{tab_main_component}
\end{table}

\begin{table}
	\centering
	\caption{The effects of different top-down maps on Flickr30K Entities test. ``VA'' means visual attention maps, while ``QA'' means our query-aware visual attention maps.}
	\scalebox{1.0}{
		\begin{tabular}{lc}
			\toprule
			Method          & Accuracy (\%) \\
			\midrule
			COM                          & 57.60\\
			\hline
			QAM (VA) + COM        & 57.85\\
			QAM (QA) + COM                  & 58.43\\	
			\hline			
			QAM (VA) + COM + SF & 57.97\\
			QAM (QA) + COM + SF & 60.15\\			
			\bottomrule
	\end{tabular}}
	\label{tab_QAM}
\end{table}

\textbf{Effects of main components.} We first study the effects of each main component in our framework (see Table \ref{tab_main_component}). Compared with the baseline method, our COM module yields 4.35\% improvements, because COM directly analyzes multi-modal data and thus avoids information loss during modality conversion. The performance of our QAM module is lower than the baseline. The reason is that our QAM can localize objects from a top-down perspective but cannot always capture accurate object boundaries. However, when we add the bounding box generated by QAM to COM (i.e., QAM+COM), we achieve higher accuracy than both the baseline and original COM, indicating that our top-down QAM and bottom-up COM are complementary. QAM can find objects which are missed in pre-extracted proposals in COM, while COM is able to select bounding boxes with better boundaries. Our SF module further fuses the top-down query-aware visual attention maps with bottom-up similarity scores, and thus improves the accuracy by 1.72\%. The learnable KAM module achieves improvements of 1.31\%, thanks to the knowledge adaptation.

\textbf{BiCM with or without training.} The QAM, COM and SF modules in our BiCM do not need any training. Compared with the baseline, our method without training yields 6.90\% improvements. Our KAM module extracts pseudo labels from unpaired training data to train an MLP. Our method with KAM outperforms the baseline by 8.21\%. 

\textbf{Different attention maps in the top-down module.} We show the effects of our query-aware visual attention maps in Table \ref{tab_QAM}. Vanilla visual attention maps are composed by attention scores $\{att_{h, u}\}_{u=1}^{U}$, which is able to highlight visually salient regions but cannot highlight regions corresponding to the language query. Therefore, it only slightly improves the performance, compared with only using COM. Our query-aware visual attention maps find out not only visual salient regions but also query-specific regions, as shown in Fig. \ref{fig_results_referit}. Thus, our query-aware visual attention maps achieve significant improvements.

\begin{figure*}
	\centerline{\includegraphics[scale=0.45]{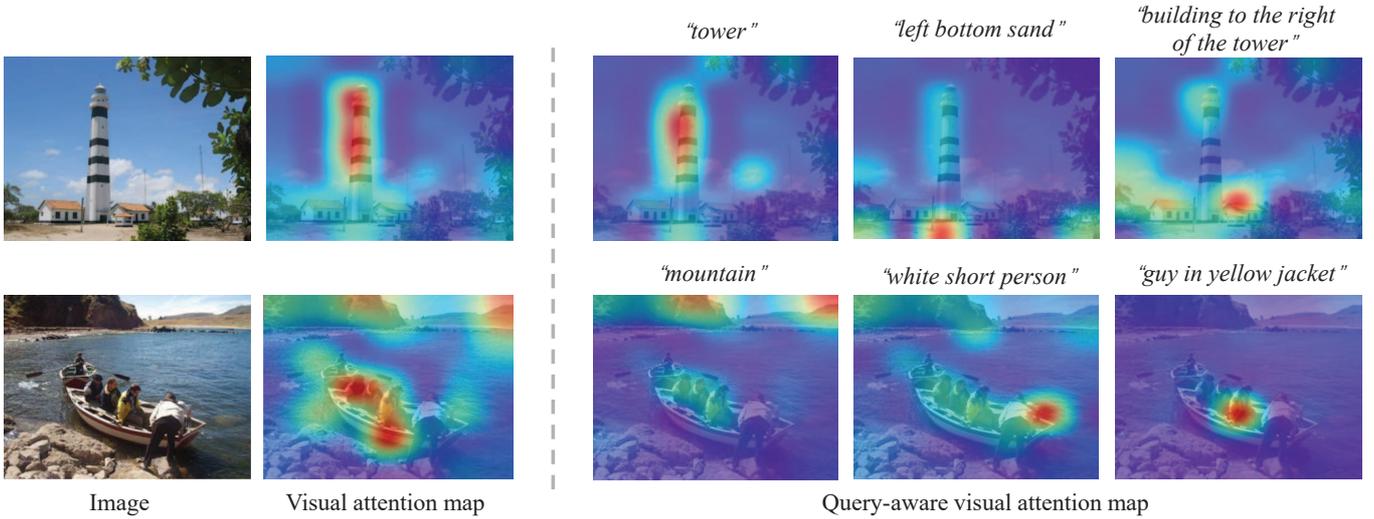}}
	\caption{Comparison of visual attention map and our proposed query-aware visual attention map on the ReferItGame dataset. }
	\label{fig_qam}
\end{figure*}

\textbf{Visualization of query-aware visual attention map.} We visualize our query-aware visual attention map in Fig.~\ref{fig_qam}. It can be seen that vanilla visual attention maps only find out visually salient regions (such as the tower in the first image in Fig.~\ref{fig_qam}), while our query-aware maps highlight objects corresponding to different language queries.

\textbf{Different information in the bottom-up module.} Table \ref{tab_COM} shows the effects of different information in our COM module. When COM only uses object class names, the performance is higher than the baseline. This is because our textual encoder models the information of the entire query while the baseline method only separately models information of every words.
Compared with using object class names, using object visual information shows a lower accuracy. A reason is that class names and queries are in the same modality while visual information and queries are in different modalities. Even though our COM embeds information in different modalities into a same space, comparing single-modal information is still easier than comparing multi-modal information. In addition, many queries only contain one word or several words, which are very similar to class names. Therefore, using class names shows better performance.
However, leveraging both class names and visual information obtains the best performance and gains significant improvements (3.51\%) compared with using only a single modality. It is because class names lack some important information such as color, posture and so on, which can be provided by vision.

\begin{table}
	\centering
	\caption{The effects of different bottom-up information settings on Flickr30K Entities test. We only use COM in this experiment. }
	\scalebox{1.0}{
		\begin{tabular}{lc}
			\toprule
			Information                             & Accuracy (\%) \\
			\midrule
			Baseline \cite{wang2019phrase}            & 53.25\\
			object class name                         & 54.09\\			
			visual object                             & 40.29\\
			visual object + class name                & 57.60\\			
			\bottomrule
	\end{tabular}}
	\label{tab_COM}
\end{table}

\textbf{Precision of pseudo labels.} To train the MLP in KAM, we extract pseudo labels from unpaired training data. Table \ref{tab_anno_thr} shows the precision of our pseudo labels, which is the percentage of correct ones in all pseudo labels we extracted. It can be observed that our precision is 74.7\% when the threshold $Thr_{k}$ is 0.9. Under this threshold, we can generate 3,986 pseudo annotations. 

\begin{table}
	\centering
	\caption{The effects of different pseudo label thresholds $Thr_{k}$ on Flickr30K Entities trainval.}
	\scalebox{1.0}{
		\begin{tabular}{ccc}
			\toprule
			& Number of  &\\
			$Thr_{k}$     & pseudo labels   & Precision (\%) \\
			\midrule
			0.6               & 6058               & 64.2    \\
			0.7               & 5180               & 68.5    \\
			0.8               & 4401               & 70.4 \\
			0.9               & 3986               & 74.7 \\
			\bottomrule
	\end{tabular}}
	\label{tab_anno_thr}
\end{table}

\begin{figure}
	\centerline{\includegraphics[scale=0.48]{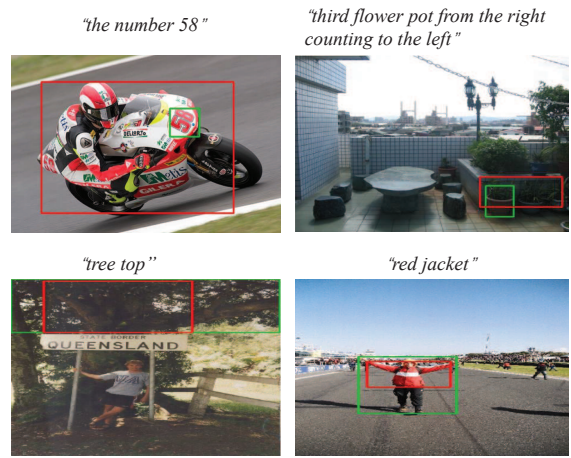}}
	\caption{Failure cases of our method. Red and green boxes are grounding results and ground truths, respectively.}
	\label{fig_failures}
\end{figure}

\textbf{Failure cases.} We depict some failure cases in Fig. \ref{fig_failures}. There are two main types of failures. The first is caused by complex reasoning. For instance, in the top two images in Fig. \ref{fig_failures}, the queries require counting and analyzing numbers, which is hard to be learned by unpaired training data. The second type is inaccurate object boundary, such as the bottom two images in Fig. \ref{fig_failures}. Our BiCM finds out the desired objects but sometimes fails to capture their boundaries. Using better object detection models can reduce this type of errors.

\section{Conclusion}
In this paper, we have presented a BiCM framework for unpaired referring grounding. It includes four major components: a top-down QAM module to extract query-aware attention maps, a bottom-up COM module to directly compare multi-modal information, an SF module to integrate top-down and bottom-up results, and a KAM module that leverages pseudo training data to adapt external knowledge to the target grounding data. Experimental results have demonstrated that our proposed method outperforms the existing state-of-the-art methods by a large margin on two popular referring grounding datasets. 

\bibliographystyle{elsarticle-num}
\bibliography{my_reference}

\end{document}